\pdfoutput=1

\documentclass[11pt]{article}

\usepackage[]{acl}

\usepackage{times}
\usepackage{latexsym}

\usepackage[T1]{fontenc}

\usepackage[utf8]{inputenc}

\usepackage{microtype}

\usepackage{tabularx}
\newcolumntype{Y}{>{\centering\arraybackslash}X}

\usepackage{graphicx}
\usepackage{booktabs,caption}

\usepackage[T1]{fontenc}
\usepackage[utf8]{inputenc}
\usepackage{booktabs} 
\usepackage{color}
\usepackage{balance}
\usepackage{amsfonts}
\usepackage{amssymb, bbm}
\usepackage{multirow, multicol}
\usepackage{caption, subcaption}
\usepackage{comment}
\usepackage{xspace}
\usepackage{upquote}
\usepackage{float}
\usepackage{alltt}
\usepackage{algpseudocode}
\usepackage{pbox}
\usepackage{enumitem}
\usepackage{mathpartir}
\usepackage{wrapfig}
\usepackage{hhline}
\usepackage{stmaryrd}
\usepackage{pifont}
\usepackage{makecell}
\usepackage{graphicx}
\usepackage{amsmath}
\usepackage{array}
\usepackage{ccicons}
\usepackage{listings}
\usepackage{url}

\title{KQA Pro: A Dataset with Explicit Compositional Programs for Complex Question Answering over Knowledge Base}

  \newcommand*{\email}[1]{\texttt{#1}}

  \author{
    \textbf{Shulin Cao}$^{1,2}\thanks{\quad indicates equal contribution}$, \textbf{Jiaxin Shi}$^{1,3*}$, \textbf{Liangming Pan}$^{4}$, \textbf{Lunyiu Nie}$^{1}$, \textbf{Yutong Xiang}$^{5}$, \\
    \textbf{Lei Hou}$^{1,2}$\thanks{\quad Corresponding Author}\hspace{0.5em}, \textbf{Juanzi Li}$^{1,2}$, \textbf{Bin He}$^{6}$, \textbf{Hanwang Zhang}$^{7}$ \\
    $^1$Department of Computer Science and Technology, BNRist \\
    $^2$KIRC, Institute for Artificial Intelligence, Tsinghua University, Beijing 100084, China \\
    $^3$Cloud BU, Huawei Technologies, $^4$National University of Singapore, $^5$ETH Zürich\\
    $^6$Noah’s Ark Lab, Huawei Technologies, $^7$Nanyang Technological University \\
    \email{\{caosl19@mails., houlei@,lijuanzi@\}tsinghua.edu.cn}\\
    \email{shijx12@gmail.com}, \email{liangmingpan@u.nus.edu} \\
    }

\begin{document}
\maketitle
\begin{abstract}
Complex question answering over knowledge base (Complex KBQA) is challenging because it requires various compositional reasoning capabilities, such as multi-hop inference, attribute comparison, set operation. 
Existing benchmarks have some shortcomings that limit the development of Complex KBQA:
1) they only provide QA pairs without explicit reasoning processes;
2) questions are poor in diversity or scale.
To this end, we introduce KQA Pro, a dataset for Complex KBQA including \textasciitilde120K diverse natural language questions.
We introduce a compositional and interpretable programming language KoPL to represent the reasoning process of complex questions.
For each question, we provide the corresponding KoPL program and SPARQL query, so that KQA Pro serves for both KBQA and semantic parsing tasks.
Experimental results show that SOTA KBQA methods cannot achieve promising results on KQA Pro as on current datasets, which suggests that KQA Pro is challenging and Complex KBQA requires further research efforts. We also treat KQA Pro as a diagnostic dataset for testing multiple reasoning skills, conduct a thorough evaluation of existing models and discuss further directions for Complex KBQA. Our codes and datasets
can be obtained from \url{https://github.com/shijx12/KQAPro_Baselines}.

\end{abstract}

\section{Introduction}

Thanks to the recent advances in deep models, especially large-scale unsupervised representation learning~\cite{devlin2019bert}, question answering of simple questions over knowledge base (Simple KBQA), i.e., single-relation factoid questions~\cite{bordes2015simplequestions}, begins to saturate~\cite{petrochuk2018simplequestions,wu2019learning,huang2019knowledge}.
However, tackling complex questions (Complex KBQA) is still an ongoing challenge, due to the unsatisfied capability of compositional reasoning. 
As shown in Table~\ref{tab:datasets}, to promote the community development, several benchmarks are proposed for Complex KBQA, including LC-QuAD2.0~\cite{dubey2019lcquad}, ComplexWebQuestions~\cite{talmor2018web}, MetaQA~\cite{zhang2018variational}, CSQA~\cite{saha2018complex}, CFQ~\cite{keysers2020measuring}, and so on.
However, they suffer from the following problems:

1) Most of them only provide QA pairs without explicit reasoning processes, making it challenging for models to learn compositional reasoning. 
Some researchers try to learn the reasoning processes with reinforcement learning~\cite{liang2017neural,saha2019complex,ansari2019neural} and searching~\cite{guo2018dialog}.
However, the prohibitively huge search space hinders both the performance and speed, especially when the question complexity increases.
For example, \citet{saha2019complex} achieved a 96.52\% F1 score on simple questions in CSQA, whereas only 0.33\% on complex questions that require comparative count.
We think that intermediate supervision is needed for learning the compositional reasoning, mimicking the learning process of human beings~\cite{holt2017children}.

2) Questions are not satisfactory in diversity and scale. For example, MetaQA~\cite{zhang2018variational} questions are generated using just 36 templates, and they only consider relations between entities, ignoring literal attributes; LC-QuAD2.0~\cite{dubey2019lcquad} and ComplexWebQuestions~\cite{talmor2018web} have fluent and diverse human-written questions, but their scale is less than 40K.

\begin{figure}[t]
  \centering
  \includegraphics[width=0.9\linewidth]{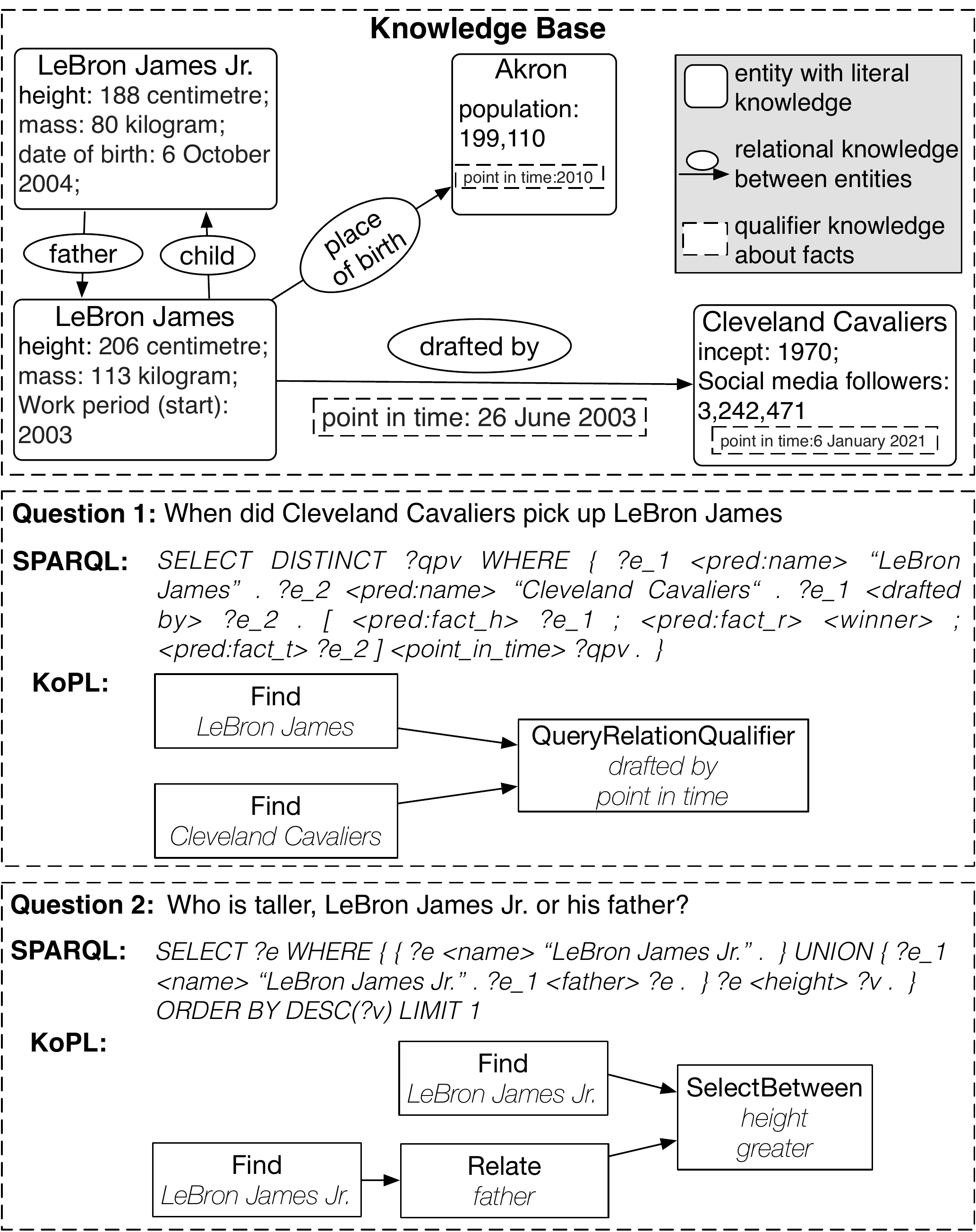}
  \caption{Example of our KB and questions. Our KB is a dense subset of Wikidata~\cite{Wikidata}, including multiple types of knowledge. Our questions are paired with executable KoPL programs and SPARQL queries.}
  \label{fig:intro}
\end{figure}

To address these problems, we create \textbf{KQA Pro}, a large-scale benchmark for Complex KBQA.
In KQA Pro, we define a Knowledge-oriented Programming Language (\textbf{KoPL}) to explicitly describe the reasoning process for solving complex questions (see Fig.~\ref{fig:intro}).
A program is composed of symbolic \textbf{functions}, which define the basic, atomic operations on KBs. The composition of functions well captures the language compositionality~\cite{Baroni2019LinguisticGA}.
Besides KoPL, following previous works~\cite{yih2016value,su2016generating}, we also provide the corresponding SPARQL for each question, which solves a complex question by parsing it into a query graph.
Compared with SPARQL, KoPL 1) provides a more explicit reasoning process. It divides the question into multiple steps, making human understanding easier and the intermediate results more transparent; 2) allows humans to control the model behavior better, potentially supporting human-in-the-loop. When the system gives a wrong answer, users can quickly locate the error by checking the outputs of intermediate functions.
We believe the compositionality of KoPL and the graph structure of SPARQL are two complementary directions for Complex KBQA.



To ensure the diversity and scale of KQA Pro, we follow the synthesizing and paraphrasing pipeline in the literature~\cite{overnight, Dual}, first synthesize large-scale (canonical question, KoPL, SPARQL) triples, and then paraphrase the canonical questions to natural language questions (NLQs) via crowdsourcing. We combine the following two factors to achieve diversity in questions: (1) To increase structural variety, we leverage a varied set of templates to cover all the possible queries through random sampling and recursive composing; (2) To increase linguistic variety, we filter the paraphases based on their edit distance with the canonical utterance. Finally, KQA Pro consists of 117,970 diverse questions that involve varied reasoning skills (\textit{e.g.}, multi-hop reasoning, value comparisons, set operations, \textit{etc.}). 
Besides a QA dataset, it also serves as a semantic parsing dataset. To the best of our knowledge, KQA Pro is currently the largest corpus for NLQ-to-SPARQL and NLQ-to-Program tasks.

We reproduce the state-of-the-art KBQA models and thoroughly evaluate them on KQA Pro. From the experimental results, we observe significant performance drops of these models compared with on existing KBQA benchmarks. It indicates that Complex KBQA is still challenging, and KQA Pro could support further explorations. We also treat KQA Pro as a diagnostic dataset for analyzing a model's capability of multiple reasoning skills, and discover weaknesses that are not widely known, \textit{e.g.}, current models struggle on comparisonal reasoning for lacking of literal knowledge (\textit{i.e.}, \textit{(LeBron James, height, 206 centimetre)}), or perform poorly on questions whose answers are not obverseved in the training set. We hope all contents of KQA Pro could encourage the community to make further breakthroughs.

\begin{table*}[ht]
\centering
\scriptsize
\begin{tabular}{cccccc}
\toprule
\multicolumn{1}{c}{Dataset} & \begin{tabular}[c]{@{}c@{}}multiple kinds\\ of knowledge\end{tabular} & \begin{tabular}[c]{@{}c@{}}number of \\ questions\end{tabular} & \begin{tabular}[c]{@{}c@{}}natural\\ language\end{tabular} & \begin{tabular}[c]{@{}c@{}}query\\graphs\end{tabular} & \begin{tabular}[c]{@{}c@{}}multi-step \\programs \end{tabular} \\ 
\midrule
WebQuestions~\cite{berant2013semantic} & $\checkmark$     &  5,810   & $\checkmark$   & $\times$  & $\times$ \\
WebQuestionSP~\cite{yih2016value}  & $\checkmark$     &  4,737   & $\checkmark$   & $\checkmark$  & $\times$ \\
GraphQuestions~\cite{su2016generating}  & $\checkmark$     & 5,166   & $\checkmark$   & $\checkmark$  & $\times$ \\
LC-QuAD2.0~\cite{dubey2019lcquad}  & $\checkmark$     & 30,000   & $\checkmark$   & $\checkmark$  & $\times$ \\
ComplexWebQuestions~\cite{talmor2018web}  & $\checkmark$ & 34,689   & $\checkmark$ & $\checkmark$   & $\times$ \\
MetaQA~\cite{zhang2018variational}  & $\times$  & 400,000  & $\times$ & $\times$  & $\times$ \\
CSQA~\cite{saha2018complex}   & $\times$  & 1.6M  & $\times$ & $\times$  & $\times$ \\
CFQ~\cite{keysers2020measuring} & $\times$  &  239,357  &  $\times$  &  $\checkmark$  &  $\times$  \\
GrailQA~\cite{GrailQA} & $\checkmark$ & 64,331 & $\checkmark$ & $\checkmark$ & $\times$ \\
\midrule
\textbf{KQA Pro (ours)}  & $\checkmark$  & 117,970  & $\checkmark$  & $\checkmark$  & $\checkmark$ \\ 
\bottomrule
\end{tabular}
\caption{\label{tab:datasets}Comparison with existing datasets of Complex KBQA. The column \textit{multiple kinds of knowledge} means whether the dataset considers multiple types of knowledge or just relational knowledge (introduced in Sec.\ref{sec:kb_definition}). The column \textit{natural language} means whether the questions are in natural language or written by templates.}
\end{table*}

\section{Related Work}

Complex KBQA aims at answering complex questions over KBs, which requires multiple reasoning capabilities such as multi-hop inference, quantitative comparison, and set operation~\cite{Lan2021ASO}. Current methods for Complex KBQA can be grouped into two categories: 1) semantic parsing based methods~\cite{liang2017neural,guo2018dialog,saha2019complex,ansari2019neural}, which parses a question to a symbolic
logic form (\textit{e.g.}, $\lambda$-calculus~\cite{artzi2013semantic}, $\lambda$-DCS~\cite{liang2013lambda,pasupat2015compositional,wang2015building,pasupat2016inferring}, SQL~\cite{zhong2017seq2sql}, AMR~\cite{banarescu2013abstract}, SPARQL~\cite{sun2020sparqa}, and \textit{etc.}) and then executes it against the KB and obtains
the final answers; 2) information retrieval based methods~\cite{miller2016key, saxena2020improving, schlichtkrull2018rgcn,zhang2018variational,zhou2018interpretable,qiu2020stepwise,Shi2021TransferNetAE}, which constructs a question-specific graph extracted from the KB and ranks all the entities in the extracted graph based on their relevance to the question. 

Compared with information retrieval based methods, semantic parsing based methods provides better interpretability by generating expressive logic forms, which represents the intermediate reasoning process. However, 
manually annotating logic forms
is expensive and labor-intensive, and it is challenging
to train a semantic parsing model with weak supervision signals
(\textit{i.e.}, question-answer pairs). Lacking logic form annotations turns out to be one of the main bottlenecks of semantic parsing. 

Table~\ref{tab:datasets} lists the widely-used datasets in Complex KBQA community and their features.
MetaQA and CSQA have a large number of questions, but they ignore literal knowledge, lack logic form annotations, and their questions are written by templates.
Query graphs (\textit{e.g.}, SPARQLs) are provided in some datasets to help solve complex questions.
However, SPARQL is weak in describing the intermediate procedure of the solution, and the scale of existing question-to-SPARQL datasets is small.



In this paper, we introduce a novel logic form KoPL, which models the procedure of Complex KBQA as a multi-step program, and provides a more explicit reasoning process compared with query graphs. Furthermore, we propose KQA Pro, a large-scale semantic parsing dataset for Complex KBQA, which contains \textasciitilde120k diverse natural language questions with both KoPL and SPARQL annotations. It is the largest NLQ-to-SPARQL dataset as far as we know. Compared with these existing datasets, KQA Pro serves as a more well-rounded benchmark.

\section{Background}
\subsection{KB Definition}
\label{sec:kb_definition}
Typically, a \textbf{KB} (\textit{e.g.}, Wikidata~\cite{Wikidata}) consists of:

\noindent\textbf{Entity}, the most basic item in KB.

\noindent\textbf{Concept}, the abstraction of a set of entities, \textit{e.g.}, \textit{basketball player}.

\noindent\textbf{Relation}, the link between entities or concepts. Entities are linked to concepts via the relation \textit{instance of}. Concepts are organized into a tree structure via relation \textit{subclass of}.

\noindent\textbf{Attribute}, the literal information of an entity. An attribute has a key and a value, which is one of four types\footnote{Wikidata also has other types like geographical and time. We omit them for simplicity and leave them for future work.}: string, number, date, and year. The number value has an extra unit, \textit{e.g.}, 206 centimetre.

\noindent\textbf{Relational knowledge}, the triple with form (entity, relation, entity), \textit{e.g.}, \textit{(LeBron James Jr., father, LeBron James)}.

\noindent\textbf{Literal knowledge}, the triple with form (entity, attribute key, attribute value), \textit{e.g.}, \textit{(LeBron James, height, 206 centimetre)}.

\noindent\textbf{Qualifier knowledge}, the triple whose head is a relational or literal triple, \textit{e.g.}, \textit{((LeBron James, drafted by, Cleveland Cavaliers), point in time, 2003)}. A qualifier also has a key and a value.

\subsection{KoPL Design}

We design KoPL, a compositional and interpretable programming language to represent the reasoning process of complex questions.
It models the complex procedure of question answering with a program of intermediate steps. Each step involves a function with a fixed number of arguments.
Every program can be denoted as a binary tree. As shown in Fig.~\ref{fig:intro}, a directed edge between two nodes represents the dependency relationship between two functions. That is, the destination function takes the output of the source function as its argument.
The tree-structured program can also be serialized by post-order traversal, and formalized as a sequence with \(n\) functions. The general form is shown below. Each function \(f_i\) takes in a list of textual arguments $\textbf{a}_{i}$, which need to be inferred according to the question, and a list of functional arguments $\textbf{b}_{i}$, which come from the output of previous functions. 
\begin{equation}
f_1(\textbf{a}_1, \textbf{b}_1) f_2(\textbf{a}_2, \textbf{b}_2)  ... f_n(\textbf{a}_n, \textbf{b}_n) 
\end{equation}
Take function \textit{Relate} as an example, it has two textual inputs: relation and direction (i.e., \textit{forward} or \textit{backward}, meaning the output is object or subject). It has one functional input: a unique entity. Its output is a set of entities that hold the specific relation with the input entity. For example, in Question 2 of Fig.~\ref{fig:intro}, the function \textit{Relate}([\textit{father, forward}], [\textit{LeBron James Jr.}]) returns LeBron James, the father of LeBron James Jr. (the direction is omitted in the figure for simplicity).



We analyze the generic, basic operations for Complex KBQA, and design 27 functions\footnote{The complete function instructions are in Appendix~\ref{sec:function in appendix}.} in KoPL. They support KB item manipulation (\textit{e.g.}, \textit{Find}, \textit{Relate}, \textit{FilterConcept}, \textit{QueryRelationQualifier}, \textit{etc.}), various reasoning skills (\textit{e.g.}, \textit{And}, \textit{Or}, \textit{etc.}), and multiple question types (\textit{e.g.}, \textit{QueryName}, \textit{SelectBetween}, \textit{etc.}).
By composing the finite functions into a KoPL program\footnote{The grammar rules of KoPL are in Appendix~\ref{apdx:grammar}.}, we can model the reasoning process of infinite complex questions. 

Note that qualifiers play an essential role in disambiguating or restricting the validity of a fact~\cite{qualifier, Liu2021RoleAwareMF}. However, they have not been adequately modeled in current KBQA models or datasets. As far as we know, we are the first to explicitly model qualifiers in Complex KBQA.

\section{KQA Pro Construction}
To build KQA Pro dataset, first, we extract a knowledge base with multiple kinds of knowledge (Section~\ref{sec:KB_Extraction}). Then, we generate canonical questions, corresponding KoPL programs and SPARQL queries with novel compositional strategies (Section~\ref{sec:strategy}). In this stage, we aim to cover all the possible queries through random sampling and recursive composing. Finally, we rewrite canonical questions into natural language via crowdsourcing (Section~\ref{sec:paraphrasing}). To further increase linguistic variety, we reject the paraphrases whose edit distance with the canonical question is small.

\subsection{Knowledge Base Extraction}
\label{sec:KB_Extraction}
We took the entities of FB15k-237~\cite{FB15K-237} as seeds, and aligned them with Wikidata via Freebase IDs\footnote{The detailed extracting process is in Appendix~\ref{apdx:KB extraction}.}. The reasons are as follows: 1) The vast amount of knowledge in the full knowledge base (\textit{e.g.}, full Freebase~\cite{Freebase} or Wikidata contains millions of entities) may cause both time and space issues, while most of the entities may never be used in questions.
2) FB15k-237 is a high-quality, dense subset of Freebase, whose alignment to Wikidata produces a knowledge base with rich literal and qualifier knowledge. We added 3,000 other entities with the same name as one of FB15k-237 entities to increase the disambiguation difficulty. The statistics of our final knowledge base are listed in Table~\ref{tab:kb_statistics}.

\begin{table}[ht]
  \centering
  \small
  \begin{tabular}{ccccc}
  \hline
  \# Con. & \# Ent. & \# Name & \# Pred. & \# Attr. \\
  \hline
  794 & 16,960 &  14,471 & 363 & 846\\
  \hline
  \end{tabular}

  \begin{tabular}{ccc}
  \hline
  \# Relational facts & \# Literal facts & \# High-level facts \\
  \hline
  415,334 & 174,539 &  309,407\\
  \hline
  \end{tabular}
  \caption{\label{tab:kb_statistics} Statistics of our knowledge base. The top lists the numbers of concepts, entities, unique entity names, predicates, and attributes. The bottom lists the numbers of different types of knowledge.}
\end{table}


\begin{figure}[t]
  \centering
  \includegraphics[width=0.95\linewidth]{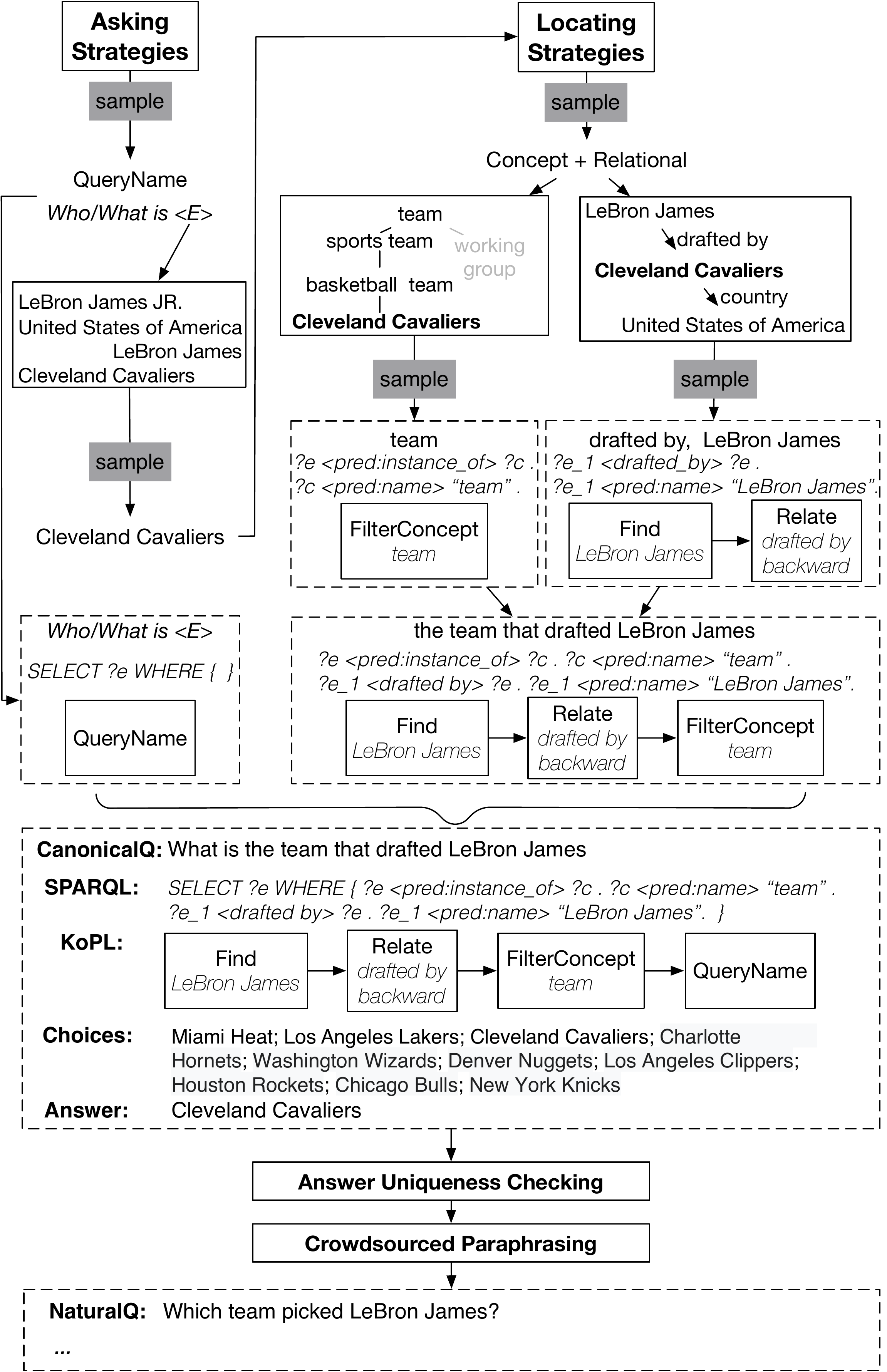}
  \caption{Process of our question generation. First, we sample a question type from asking strategies and sample a target entity from KB. Next, we sample a locating strategy and detailed conditions to describe the target entity. Finally, we combine intermediate snippets into the complete question and check whether the answer is unique. \textbf{Note that the snippets of canonical question, SPARQL, and KoPL are operated simultaneously.}. A more detailed explanation of this example is in Appendix~\ref{apdx:generation examples}.}
  \label{fig:generation}
\end{figure}

\begin{table*}[ht]
  \centering
  \scriptsize
  \begin{tabularx}{\textwidth}{l *{3}{Y}}
  \toprule
  \textbf{Strategy}  &  \textbf{Template}  &  \textbf{Example} \\
  \midrule
  \multicolumn{3}{c}{\textbf{Locating Stage}} \\
  \midrule
  Entity Name  &   -   &   LeBron James \\
  \cmidrule(lr){1-3}
  Concept + Literal  &   the \texttt{<C>} whose \texttt{<K>} is \texttt{<OP>} \texttt{<V>} (\texttt{<QK>} is \texttt{<QV>})  &   the basketball team whose social media followers is greater than 3,000,000 (point in time is 2021) \\
  \cmidrule(lr){1-3}
  Concept + Relational  &  the \texttt{<C>} that \texttt{<P>} \texttt{<E>} (\texttt{<QK>} is \texttt{<QV>})  &   the basketball player that was drafted by Cleveland Cavaliers \\
  \cmidrule(lr){1-3}
  Recursive Multi-Hop  &   unfold \texttt{<E>} in a \textit{Concept + Relational} description   &   the basketball player that was drafted by the basketball team whose social media followers is greater than 3,000,000 (point in time is 2021) \\
  \cmidrule(lr){1-3}
  Intersection   &  Condition 1 \textit{and} Condition 2  &   the basketball players whose height is greater than 190 centimetres and less than 220 centimetres \\
  \midrule
  \multicolumn{3}{c}{\textbf{Asking Stage}} \\
  \midrule
  QueryName    &    What/Who is \texttt{<E>}    &    Who is the basketball player whose height is equal to 206 centimetres? \\
  \cmidrule(lr){1-3}
  Count    &    How many \texttt{<E>}     &     How many basketball players that were drafted by Cleveland Cavaliers? \\
  \cmidrule(lr){1-3}
  SelectAmong    &    Among \texttt{<E>}, which one has the largest/smallest \texttt{<K>}    &   Among basketball players, which one has the largest mass? \\
  \cmidrule(lr){1-3}
  Verify   &   For \texttt{<E>}, is his/her/its \texttt{<K>} \texttt{<OP>} \texttt{<V>} (\texttt{<QK>} is \texttt{<QV>})   &  For the human that is the father of LeBron James Jr., is his/her height greater than 180 centimetres? \\
  \cmidrule(lr){1-3}
  QualifierRelational   &   \texttt{<E>} \texttt{<P>} \texttt{<E>}, what is the \texttt{<QK>}   &   LeBron James was drafted by Cleveland Cavaliers, what is the point in time?  \\
  \bottomrule
  \end{tabularx}
  \caption{\label{tab:example_templates} Representative templates and examples of our locating and asking stage. Placeholders in template have specific implication: \texttt{<E>}-description of an entity or entity set; \texttt{<C>}-concept; \texttt{<K>}-attribute key; \texttt{<OP>}-operator, selected from \{=, !=, $<$, $>$\}; \texttt{<V>}-attribute value; \texttt{<QK>}-qualifier key; \texttt{<QV>}-qualifier value; \texttt{<P>}-relation description, \textit{e.g.}, \textit{was drafted by}. The complete instruction is in Appendix~\ref{apdx:generation strategies}.}
\end{table*}

\subsection{Question Generation Strategies}\label{sec:strategy}
To generate diverse complex questions in a scalable manner, we propose to divide the generation into two stages: \textbf{locating} and \textbf{asking}.
In locating stage we describe a single entity or an entity set with various restrictions, while in asking stage we query specific information about the target entity or entity set.
We define several strategies for each stage.
By sampling from them and composing the two stages, we can generate large-scale and diverse questions with a small number of templates.
Fig.~\ref{fig:generation} gives an example of our generation process.

For locating stage, we propose 7 strategies and show part of them in the top section of Table~\ref{tab:example_templates}.
We can fill the placeholders of templates by sampling from KB to describe a target entity.
We support quantitative comparisons of 4 operations: equal, not equal, less than, and greater than, indicated by ``\texttt{<OP>}'' of the template.
We support optional qualifier restrictions, indicated by ``(\texttt{<QK>} is \texttt{<QV>})'', which can narrow the located entity set.
In \textit{Recursive Multi-Hop}, we replace the entity of a relational condition with a more detailed description, so that we can easily increase the hop of questions.
For asking stage, we propose 9 strategies and show some of them in the bottom section of Table~\ref{tab:example_templates}.
Our \textit{SelectAmong} is similar to \textit{argmax} and \textit{argmin} operations in $\lambda$-DCS. The complete generation strategies are shown in Appendix~\ref{apdx:generation strategies} due to space limit.

Our generated instance consists of five elements: question, SPARQL query, KoPL program, 10 answer choices, and a golden answer.
Choices are selected by executing an abridged SPARQL\footnote{The SPARQL implementation details are shown in Appendix~\ref{apdx:sparql}.}, which randomly drops one clause from the complete SPARQL.
With these choices, KQA Pro supports both multiple-choice setting and open-ended setting.
We randomly generate lots of questions, and only preserve those with a unique answer.
For example, since \textit{Akron} has different populations in different years, we will drop questions like \textit{What is the population of Akron}, unless the time constraint (\textit{e.g.}, \textit{in 2010}) is specified.




\begin{figure}[ht]
  \centering
  \includegraphics[width=1.0\linewidth]{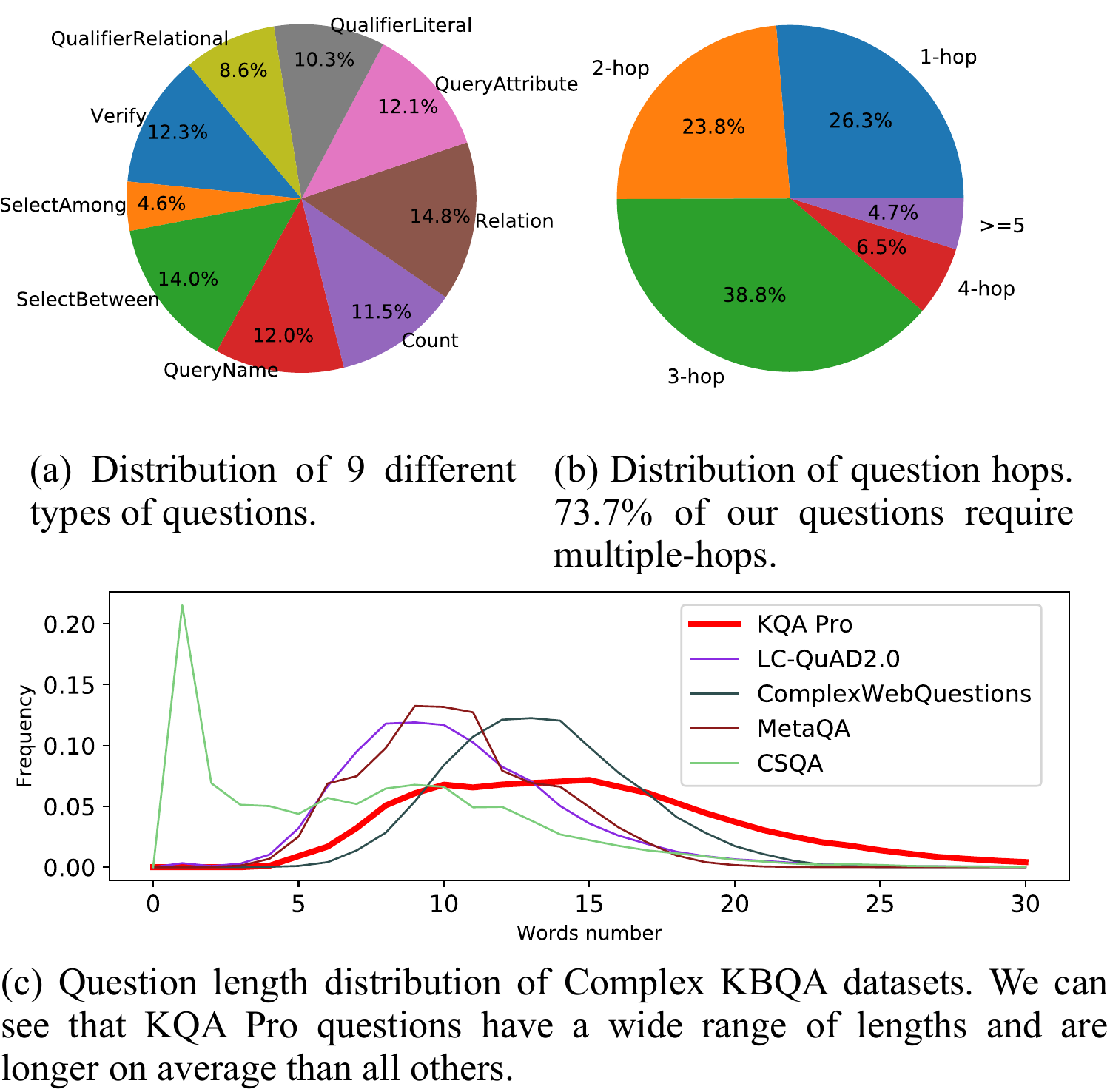}
  \caption{Question statistics of KQA Pro.}
  \label{fig:statistics}
\end{figure}

\subsection{Question Paraphrasing and Evaluation}\label{sec:paraphrasing}
After large-scale generation, we release the generated questions on Amazon Mechanical Turk (AMT) and ask the workers to paraphrase them without changing the original meaning.
For the convenience of paraphrasing, we visualize the KoPL flowcharts like Fig.~\ref{fig:intro} to help workers understand complex questions.
We allow workers to mark a question as confusing if they cannot understand it or find logical errors.
These instances will be removed from our dataset.

After paraphrasing, we evaluate the quality by 5 other workers.
They are asked to check whether the paraphrase keeps the original meaning and give a fluency rating from 1 to 5.
We reject those paraphrases which fall into one of the following cases:
(1) marked as different from the original canonical question by more than 2 workers;
(2) whose average fluency rating is lower than 3;
(3) having a very small edit distance with the canonical question.

\subsection{Dataset Analysis}

Our KQA Pro dataset consists of 117,970 instances with 24,724 unique answers.
Fig.~\ref{fig:statistics}(a) shows the question type distribution of KQA Pro.
Within the 9 types, \textit{SelectAmong} accounts for the least fraction (4.6\%), while others account for more or less than 10\%.
Fig.~\ref{fig:statistics}(b) shows that multi-hop questions cover 73.7\% of KQA Pro, and 4.7\% questions even require at least 5 hops.
We compare the question length distribution of different Complex KBQA datasets in Fig.~\ref{fig:statistics}(c).
We observe that our KQA Pro has longer questions than others on average.
In KQA Pro, the average length of questions/programs/SPARQLs is 14.95/4.79/35.52 respectively. More analysis is included in Appendix~\ref{sec:anaylisis in appendix}.


\section{Experiments}
The primary goal of our experiments is to show the challenges of KQA Pro and promising Complex KBQA directions. First, we compare the performance of state-of-the-art KBQA models on current datasets and KQA Pro, to show whether KQA Pro is challenging. Then, we treat KQA Pro as a diagnostic dataset to investigate fine-grained reasoning abilities of models, discuss current weakness and promising directions. We further conduct an experiment to explore the generation ability of our proposed model. Last, we provide a case study to show the interpretablity of KoPL.

\subsection{Experimental Settings}
\noindent \textbf{Benchmark Settings}.
We randomly split KQA Pro to train/valid/test set by 8/1/1, resulting in three sets with 94,376/11,797/11,797 instances. About 30\% answers of the test set are not seen in training.

\noindent \textbf{Representative Models}.
KBQA models typically follow a retrieve-and-rank paradigm, by constructing a question-specific graph extracted from the KB and ranks all the entities in the graph based on their relevance to the question~\cite{miller2016key, saxena2020improving, schlichtkrull2018rgcn,zhang2018variational,zhou2018interpretable,qiu2020stepwise}; or follow a parse-then-execute paradigm, by parsing a question to a query graph~\cite{berant2013semantic,yih2015semantic} or program~\cite{liang2017neural,guo2018dialog,saha2019complex,ansari2019neural} through learning from question-answer pairs.

Experimenting with all methods is logistically challenging, so we reproduce a representative subset of mothods:
\noindent \textbf{KVMemNet}~\cite{miller2016key}, a well-known model which organizes the knowledge into a memory of key-value pairs, and iteratively reads memory to update its query vector. 
\noindent \textbf{EmbedKGQA}~\cite{saxena2020improving}, a state-of-the art model on MetaQA, which incorporates knowledge embeddings to improve the reasoning performance.
\noindent \textbf{SRN}~\cite{qiu2020stepwise}, a typical path search model to start from a topic entity and predict a sequential relation path to find the target entity.
\noindent \textbf{RGCN}~\cite{schlichtkrull2018rgcn}, a variant of graph convolutional networks, tackling Complex KBQA through the natural graph structure of knowledge base.

\noindent \textbf{Our models.} Since KQA Pro provides the annotations of SPARQL and KoPL, we directly learn our parsers using supervised learning by regarding the semantic parsing as a sequence-to-sequence task.
We explore the widely-used sequence-to-sequence model---\textbf{RNN} with attention mechanism~\cite{dong2016language}, and the pretrained generative language model---\textbf{BART}~\cite{lewis2019bart}, as our SPARQL and KoPL parsers.

For KoPL learning, we design a serializer to translate the tree-structured KoPL to a sequence. For example, the KoPL program in Fig.~\ref{fig:generation} is serialized as:
\textit{Find $\langle arg \rangle$ LeBron James $\langle func \rangle$ Relate $\langle arg \rangle$ drafted by $\langle arg \rangle$ backward $\langle func \rangle$ FilterConcept $\langle arg \rangle$ team $\langle func \rangle$ QueryName}.
Here, \textit{$\langle arg \rangle$} and \textit{$\langle func \rangle$} are special tokens we designed to indicate the structure of KoPL.

To compare machine with \textbf{Human}, we sample 200 instances from the test set, and ask experts to answer them by searching our knowledge base.

\noindent \textbf{Implementation Details}.
For our BART model, we used the bart-base model of HuggingFace\footnote{https://github.com/huggingface/transformers}.
We used the optimizer Adam~\cite{kingma2014adam} for all models. 
We searched the learning rate for BART paramters in \{1e-4, 3e-5, 1e-5\}, the learning rate for other parameters in \{1e-3, 1e-4, 1e-5\}, and the weight decay in \{1e-4, 1e-5, 1e-6\}. According to the performance on validation set, we finally used learning rate 3e-5 for BART parameters, 1e-3 for other parameters, and weight decay 1e-5.


\subsection{Difficulty of KQA Pro}
We compare the performance of KBQA models on KQA Pro with MetaQA and WebQSP (short for WebQuestionSP), two commonly used benchmarks in Complex KBQA. 
The experimental results are in Table~\ref{tab:overal_results}, from which we observe that:

Although the models perform well on MetaQA and WebQSP, their performances are significantly lower and not satisfying on KQA Pro. It indicates that our KQA Pro is challenging and the Complex KBQA still needs more research efforts. Actually, 1) Both MetaQA and WebQSP mainly focus on relational knowledge, \textit{i.e.}, multi-hop questions. Therefore, previous models on these datasets are designed to handle only entities and relations. In comparison, KQA Pro includes three types of knowledge, \textit{i.e.}, relations, attributes, and qualifiers, thus is much more challenging. 2) Compared with MetaQA which contains template questions, KQA Pro contains diverse natural language questions and can evaluate models' language understanding abilities. 3) Compared with WebQSP which contains 4,737 fluent and natural questions, KQA Pro covers more question types (\textit{e.g.}, verification, counting) and reasoning operations (\textit{e.g.}, intersect, union).



\begin{table}[ht]
\centering
\scalebox{0.7}{
\begin{tabular}{lccccc}
\toprule
\textbf{Model}  &  \multicolumn{3}{c}{\textbf{MetaQA}}  &  \textbf{WebQSP}  &  \textbf{KQAPro} \\
  &  1-hop  &  2-hop   &  3-hop  &  & \\
\midrule
KVMemNet  &  96.2  &  82.7  &  48.9  &   46.7  &  16.61   \\
SRN   &  97.0   &  95.1   &   75.2   &  -  &  12.33 \\
EmbedKGQA   &  97.5   &  98.8   &   94.8  &   66.6   &  28.36 \\
RGCN    & - & - & - & 37.2  & 35.07 \\
BART & - & -& 99.9 & 67.5 & 90.55 \\
\bottomrule
\end{tabular}
}
\caption{\label{tab:overal_results} SOTA models of Complex KBQA and their performance on different datasets. SRN's result on KQA Pro, 12.33\%, is obtained on questions about only relational knowledge. The RGCN results on WebQSP is from~\cite{sun2018open}. The BART results on MetaQA 3-hop WebQSP results are from~\cite{huang2021unseen}.} 
\end{table}

\begin{table*}[!htbp]
\centering
\small

\begin{tabularx}{0.9\textwidth}{l *{8}{Y}}
\toprule
Model & \textbf{Overall} & Multi-hop & Qualifier & Compari-son & Logical & Count & Verify & Zero-shot \\ 
\midrule
KVMemNet  &  16.61  &  16.50   &  18.47  & 1.17  &  14.99  &  27.31  &  54.70  &  0.06 \\
SRN  &  -  &   12.33  &  -  &  -   &  -  &  -  &  -  & - \\
EmbedKGQA  &  28.36  &  26.41  &  25.20  &  11.93  &  23.95  &  32.88  &  61.05  &  0.06  \\
RGCN  &  35.07  &  34.00  &  27.61  &  30.03  & 35.85  & 41.91  &  65.88  &  0.00 \\
\midrule
RNN SPARQL &  41.98  &  36.01   &  19.04  &  66.98  &  37.74  & 50.26  &  58.84  &  26.08  \\
RNN KoPL  &  43.85  &  37.71  &  22.19  & 65.90  & 47.45  &  50.04 & 42.13  &  34.96  \\
BART SPARQL  &  89.68  &  88.49  &  83.09  &  96.12  &  88.67  &  85.78  &  92.33  &  87.88 \\
BART KoPL  &  90.55  &  89.46  &  84.76  &  95.51  &  89.30  &  86.68  &  93.30  &  89.59 \\
BART KoPL$^\mathbf{CG}$ & 77.86 & 77.86 & 61.46 & 93.61 & 77.88 & 79.17 & 89.01 & 76.04 \\
\midrule


Human  &  97.50  &  97.24   &  95.65  &  100.00  & 98.18  & 83.33  & 95.24  &  100.00  \\
\bottomrule
\end{tabularx}
\caption{\label{tab:in-depth results}Accuracy of different models on KQA Pro test set. BART KoPL$^\mathbf{CG}$ denotes the BART based KoPL parser on the compositional generalization experiment (see Section~\ref{sec:cg})}
\end{table*}
\subsection{Analyses on Reasoning Skills}
KQA Pro can serve as a diagnostic dataset for in-depth analyses of reasoning abilities (\textit{e.g.}, counting, comparision, logical reasoning, \textit{etc.}) for Complex KBQA, since KoPL programs underlying the questions provide tight control over the dataset.

We categorize the test questions to measure fine-grained ability of models.
Specifically, \emph{Multi-hop} means multi-hop questions, \emph{Qualifier} means questions containing qualifier knowledge, \emph{Comparison} means quantitative or temporal comparison between two or more entities, \emph{Logical} means logical union or intersection, \emph{Count} means questions that ask the number of target entities, \emph{Verify} means questions that take ``yes'' or ``no'' as the answer, \emph{Zero-shot} means questions whose answer is not seen in the training set. 
The results are shown in Table~\ref{tab:in-depth results}, from which we have the following observations:

(1) Benefits of intermediate reasoning supervision. Our RNN and BART models outperform current models significantly on all reasoning skills. This is because KoPL program and SPARQL query provide intermediate supervision which benefits the learning process a lot. As~\cite{Dua2020BenefitsOI} suggests, future dataset
collection efforts should set aside a fraction of budget for intermediate annotations, particularly as the
reasoning required becomes more complex. We hope our dataset KQA Pro with KoPL and SPARQL annotations will help guide further research in Complex KBQA.
(2) More attention to literal and qualifier knowledge. Existing models perform poorly in situations requiring comparison capability. 
This is because they only focus the relational knowledge, while ignoring the literal and qualifier knowledge. We hope our dataset will encourage the community to pay more attention to multiple kinds of knowledge in Complex KBQA.
(3) Generalization to novel questions and answers. For zero-shot questions, current models all have a close to zero performance. This indicates the models solve the questions by simply memorizing their training data, and perform poorly on generalizing to novel questions and answers. 
\subsection{Compositional Generalization}
\label{sec:cg}
We further use KQA Pro to test the ability of KBQA models to generalize to questions that contain novel combinations of
the elements observed during training.
Following previous works, we conduct the ``productivity'' experiment~\cite{Lake2018GeneralizationWS, Shaw2021CompositionalGA}, which focuses on generalization to longer sequences or to greater compositional depths than have been seen in training (for example, from a length 4 program to a length 5 program).
Specifically, we take the instances with short programs as training examples, and those with long programs as test and valid examples, resulting in three sets including 106,182/5,899/5,899 examples. The performance of BART KoPL drops from 90.55\% to 77.86\%, which indicates learning to generalize
compositionally for pretrained language models requires more research efforts. Our KQA Pro provides an environment for further experimentation on compositional generalization.


\subsection{Case Study}
\begin{figure}[t]
  \centering
  \includegraphics[width=\linewidth]{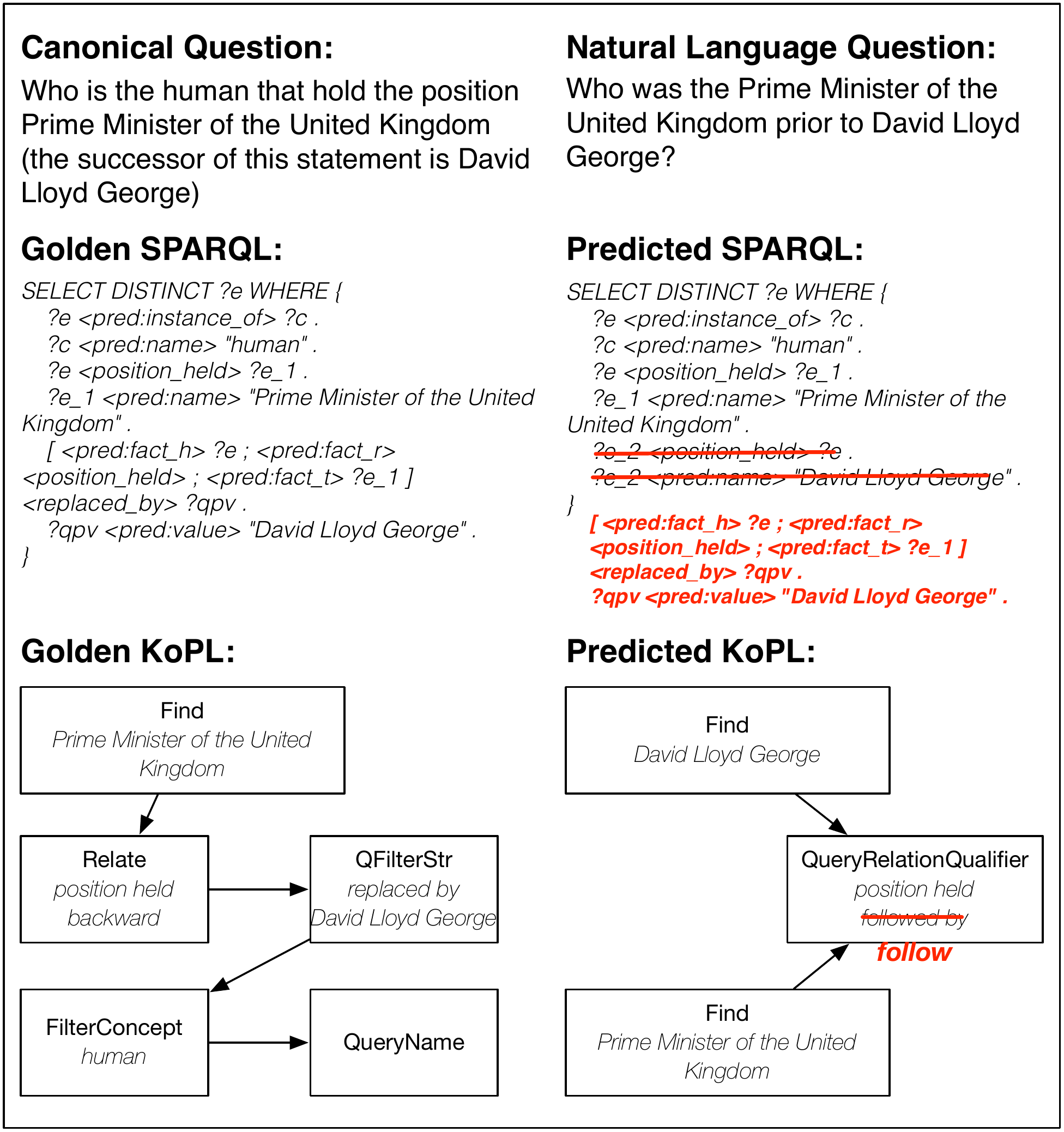}
    \caption{\label{fig:case_study}Predicted SPARQL and KoPL by BART. We show the natural language question and canonical question before human rewriting. We mark the error corrections of the wrong predictions in red.}
\end{figure}
To further understand the quality of logical forms predicted by the BART parser, we show a case in Fig.~\ref{fig:case_study}, for which the SPARQL and KoPL parsers both give wrong predictions.
The SPARQL parser fails to understand \emph{prior to David Lloyd George} and gives a totally wrong prediction for this part.
The KoPL parser gives a function prediction which is semantically correct but very different from our generated golden one.
It is a surprising result, revealing that the KoPL parser can understand the semantics and learn multiple solutions for each question, similar to the learning process of humans.
We manually correct the errors of predicted SPARQL and KoPL and mark them in red.
Compared to SPARQLs, KoPL programs are easier to be understood and more friendly to be modified.

\section{Conclusion and Future Work}
In this work, we introduce a large-scale dataset with explicit compositional programs for Complex KBQA. For each question, we provide the corresponding KoPL program and SPARQL query so that  KQA Pro can serve for both KBQA and semantic parsing tasks. We conduct a thorough evaluation of various models, discover weaknesses of current models and discuss future directions.
Among these models, the KoPL parser shows great interpretability.
As shown in Fig.~\ref{fig:case_study},
when the model predicts the answer, it will also give a reasoning process and a confidence score (which is ommited in the figure for simplicity). 
When the parser makes mistakes, humans can easily locate the error through reading the human-like reasoning process or checking the outputs of intermediate functions. In addition, using human correction data, the parser can be incrementally trained to improve the performance continuously. We will leave this as our future work.

\section*{Acknowledgments}
This work is founded by the National Key Research and Development Program of China (2020AAA0106501), the Institute for Guo Qiang, Tsinghua University (2019GQB0003), Huawei Noah’s Ark Lab, Beijing Academy of Artificial Intelligence and MOE Tier 2 funding.




\bibliography{anthology}
\bibliographystyle{acl_natbib}
\clearpage
\appendix

\begin{figure*}[ht]
  \centering
  \includegraphics[width=1.0\textwidth]{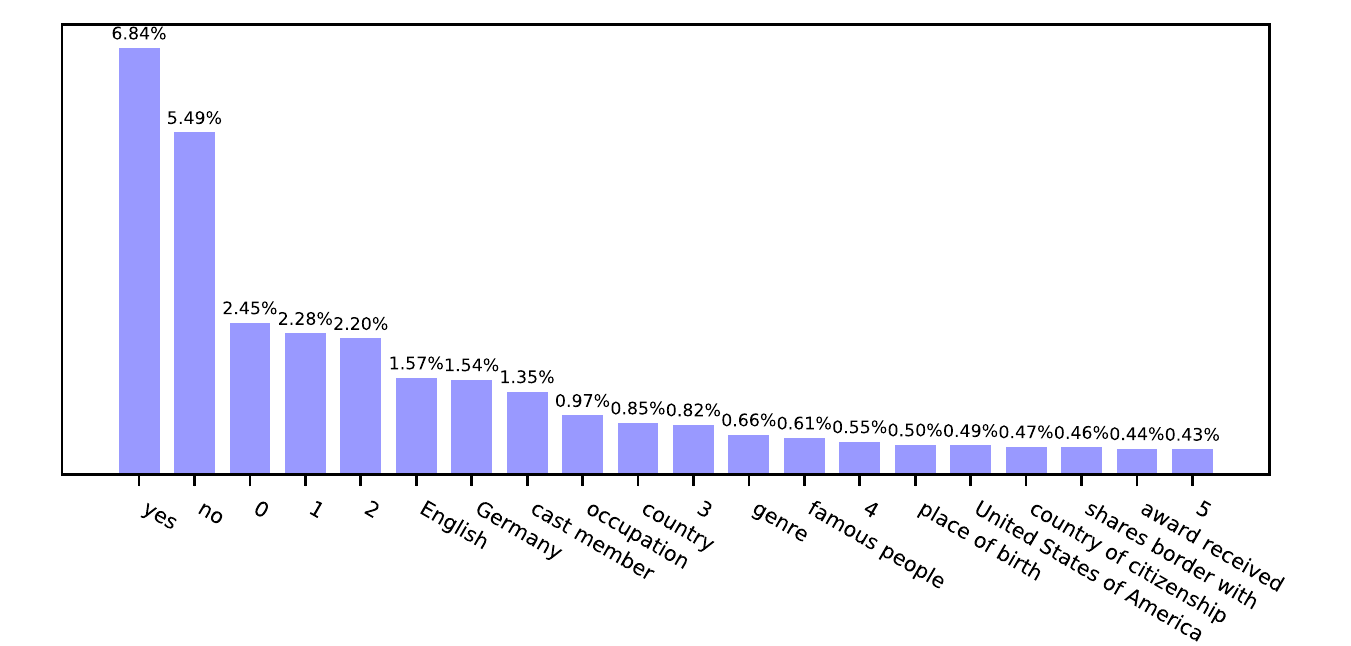}
  \caption{Top  20  most  occurring  answers  in  KQA  Pro.  The  most frequent one is ''yes``, which is the answer of about half of type \textit{Verify}.}
  \label{fig:answerstatistics}
\end{figure*}

\begin{figure}[ht]
  \centering
  \includegraphics[width=1.0\linewidth]{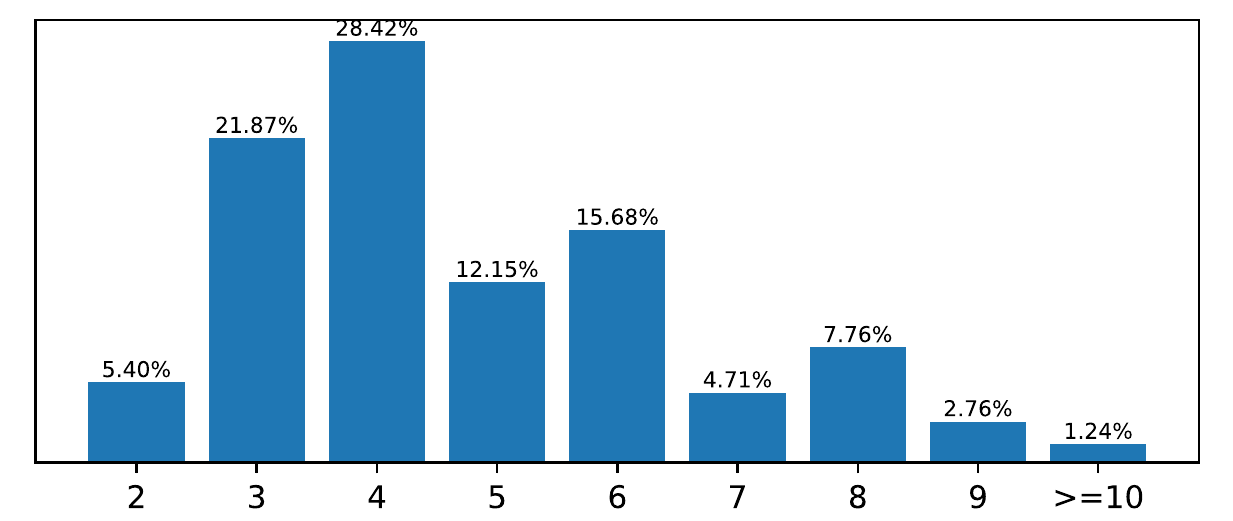}
  \caption{Distribution of program lengths.}
  \label{fig:programlenstatistics}
\end{figure}

\section{Function Library of KoPL}
\label{sec:function in appendix}
\begin{table*}[h]
\centering
\scriptsize
\begin{tabularx}{\textwidth}{l *{3}{Y}}
\toprule
\textbf{Function} & \textbf{Functional Inputs $\times$ Textual Inputs $\rightarrow$ Outputs} & \textbf{Description} & \textbf{Example} (only show textual inputs) \\
\midrule

\textit{FindAll} &   () $\times$ () $\rightarrow$ (\textit{Entities})   &  Return all entities in KB  & - \\
\hline

\textit{Find} & () $\times$ (\textit{Name}) $\rightarrow$ (\textit{Entities})  &  Return all entities with the given name  &  \textit{Find(LeBron James)} \\
\hline

\textit{FilterConcept} & (\textit{Entities}) $\times$ (\textit{Name}) $\rightarrow$ (\textit{Entities})  &  Find those belonging to the given concept  &  \textit{FilterConcept(athlete)} \\
\hline

\textit{FilterStr} & (\textit{Entities}) $\times$ (\textit{Key}, \textit{Value}) $\rightarrow$ (\textit{Entities}, \textit{Facts}) & Filter entities with an attribute condition of string type, return entities and corresponding facts  &  \textit{FilterStr(gender, male)} \\
\hline

\textit{FilterNum} & (\textit{Entities}) $\times$ (\textit{Key}, \textit{Value}, \textit{Op}) $\rightarrow$ (\textit{Entities}, \textit{Facts}) & Similar to \textit{FilterStr}, except that the attribute type is number  &  \textit{FilterNum(height, 200 centimetres, $>$)} \\
\hline

\textit{FilterYear} & (\textit{Entities}) $\times$ (\textit{Key}, \textit{Value}, \textit{Op}) $\rightarrow$ (\textit{Entities}, \textit{Facts}) & Similar to \textit{FilterStr}, except that the attribute type is year  &  \textit{FilterYear(birthday, 1980, $=$)} \\
\hline

\textit{FilterDate} & (\textit{Entities}) $\times$ (\textit{Key}, \textit{Value}, \textit{Op}) $\rightarrow$ (\textit{Entities}, \textit{Facts}) & Similar to \textit{FilterStr}, except that the attribute type is date  &  \textit{FilterDate(birthday, 1980-06-01, $<$)} \\
\hline

\textit{QFilterStr} & (\textit{Entities}, \textit{Facts}) $\times$ (\textit{QKey}, \textit{QValue}) $\rightarrow$ (\textit{Entities}, \textit{Facts}) & Filter entities and corresponding facts with a qualifier condition of string type  &  \textit{QFilterStr(language, English)} \\
\hline

\textit{QFilterNum} & (\textit{Entities}, \textit{Facts}) $\times$ (\textit{QKey}, \textit{QValue}, \textit{Op}) $\rightarrow$ (\textit{Entities}, \textit{Facts}) & Similar to \textit{QFilterStr}, except that the qualifier type is number  &  \textit{QFilterNum(bonus, 20000 dollars, $>$)} \\
\hline

\textit{QFilterYear} & (\textit{Entities}, \textit{Facts}) $\times$ (\textit{QKey}, \textit{QValue}, \textit{Op}) $\rightarrow$ (\textit{Entities}, \textit{Facts}) & Similar to \textit{QFilterStr}, except that the qualifier type is year  &  \textit{QFilterYear(start time, 1980, $=$)} \\
\hline

\textit{QFilterDate} & (\textit{Entities}, \textit{Facts}) $\times$ (\textit{QKey}, \textit{QValue}, \textit{Op}) $\rightarrow$ (\textit{Entities}, \textit{Facts}) & Similar to \textit{QFilterStr}, except that the qualifier type is date  &  \textit{QFilterDate(start time, 1980-06-01, $<$)} \\
\hline

\textit{Relate} & (\textit{Entity}) $\times$ (\textit{Pred}, \textit{Dir})
$\rightarrow$ (\textit{Entities}, \textit{Facts}) & Find entities that have a specific relation with the given entity & \textit{Relate(capital, forward)} \\
\hline

\textit{And} & (\textit{Entities}, \textit{Entities}) $\times$ () $\rightarrow$ (\textit{Entities}) & Return the intersection of two entity sets & - \\
\hline
\textit{Or} & (\textit{Entities}, \textit{Entities}) $\times$ () $\rightarrow$ (\textit{Entities}) & Return the union of two entity sets & - \\
\hline

\textit{QueryName} & (\textit{Entity}) $\times$ () $\rightarrow$ (string) & Return the entity name & - \\
\hline

\textit{Count} & (\textit{Entities}) $\times$ () $\rightarrow$ (number) & Return the number of entities & - \\
\hline

\textit{QueryAttr} & (\textit{Entity}) $\times$ (\textit{Key}) $\rightarrow$ (\textit{Value}) & Return the attribute value of the entity & \textit{QueryAttr(height)} \\
\hline

\textit{QueryAttrUnderCondition} & (\textit{Entity}) $\times$ (\textit{Key}, \textit{QKey}, \textit{QValue}) $\rightarrow$ (\textit{Value}) & Return the attribute value, whose corresponding fact should satisfy the qualifier condition & \textit{QueryAttrUnderCondition(population, point in time, 2019)} \\
\hline

\textit{QueryRelation} & (\textit{Entity}, \textit{Entity}) $\times$ ()
$\rightarrow$ (\textit{Pred}) & Return the relation between two entities & \textit{QueryRelation(LeBron James, Cleveland Cavaliers)} \\
\hline

\textit{SelectBetween} & (\textit{Entity}, \textit{Entity}) $\times$ (\textit{Key}, \textit{Op}) $\rightarrow$ (string) & From the two entities, find the one whose attribute value is greater or less and return its name & \textit{SelectBetween(height, greater)} \\
\hline

\textit{SelectAmong} & (\textit{Entities}) $\times$ (\textit{Key}, \textit{Op}) $\rightarrow$ (string) & From the entity set, find the one whose attribute value is the largest or smallest & \textit{SelectAmong(height, largest)} \\
\hline

\textit{VerifyStr} & (\textit{Value}) $\times$ (\textit{Value}) $\rightarrow$ (boolean) & Return whether the output of \textit{QueryAttr} or \textit{QueryAttrUnderCondition} and the given value are equal as string & \textit{VerifyStr(male)} \\
\hline

\textit{VerifyNum} & (\textit{Value}) $\times$ (\textit{Value}, \textit{Op}) $\rightarrow$ (boolean) & Return whether the two numbers satisfy the condition & \textit{VerifyNum(20000 dollars, $>$)} \\
\hline

\textit{VerifyYear} & (\textit{Value}) $\times$ (\textit{Value}, \textit{Op}) $\rightarrow$ (boolean) & Return whether the two years satisfy the condition & \textit{VerifyYear(1980, $>$)} \\
\hline

\textit{VerifyDate} & (\textit{Value}) $\times$ (\textit{Value}, \textit{Op}) $\rightarrow$ (boolean) & Return whether the two dates satisfy the condition & \textit{VerifyDate(1980-06-01, $>$)} \\
\hline

\textit{QueryAttrQualifier} & (\textit{Entity}) $\times$ (\textit{Key}, \textit{Value}, \textit{QKey})
$\rightarrow$ (\textit{QValue}) & Return the qualifier value of the fact (\textit{Entity}, \textit{Key}, \textit{Value}) & \textit{QueryAttrQualifier(population, 199,110, point in time)} \\
\hline

\textit{QueryRelationQualifier} & (\textit{Entity}, \textit{Entity}) $\times$ (\textit{Pred}, \textit{QKey})
$\rightarrow$ (\textit{QValue}) & Return the qualifier value of the fact (\textit{Entity}, \textit{Pred}, \textit{Entity}) & \textit{QueryRelationQualifier(drafted by, point in time)} \\

\bottomrule
\end{tabularx}
\caption{\label{tab:functions}Details of our 27 functions. Each function has 2 kinds of inputs: the functional inputs come from the output of previous functions, while the textual inputs come from the question.}
\end{table*}

Table~\ref{tab:functions} shows our 27 functions and their explanations.
Note that we define specific functions for different attribute types (i.e., string, number, date, and year), because the comparison of these types are quite different.
Following we explain some necessary items in our functions.

\noindent\textit{Entities/Entity}: \textit{Entities} denotes an entity set, which can be the output or functional input of a function. When the set has a unique element, we get an \textit{Entity}.

\noindent\textit{Name}: A string that denotes the name of an entity or a concept.

\noindent \textit{Key/Value}: The key and value of an attribute.

\noindent \textit{Op}: The comparative operation. It is one of $\{=,\ne,<,>\}$ when comparing two values, one of $\{\textit{greater}, \textit{less}\}$ in \textit{SelectBetween}, and one of $\{\textit{largest}, \textit{smallest}\}$ in \textit{SelectAmong}.

\noindent\textit{Pred/Dir}: The relation and direction of a relation.

\noindent\textit{Fact}: A literal fact, \textit{e.g.}, (\textit{LeBron James}, \textit{height}, \textit{206 centimetre}), or a relational fact, \textit{e.g.}, (\textit{LeBron James}, \textit{drafted by}, \textit{Cleveland Cavaliers}).

\noindent\textit{QKey/QValue}: The key and value of a qualifier.

\begin{table*}[t!]
\centering
\small
\resizebox{\textwidth}{!}{
\begin{tabular}{p{0.18\textwidth}>{\raggedright}p{0.45\textwidth}>{\raggedright\arraybackslash}p{0.4\textwidth}}
\toprule
\textbf{Non-terminal} & \textbf{KoPL Program} & \textbf{Canonical Question} \\
\midrule
    \texttt{<ROOT>} $\rightarrow$ & \texttt{<ES>} \textit{QueryName()} & [ What | Who ] is \texttt{<ES>} \\
    & \texttt{<ES>} \textit{Count()}  & How many \texttt{<ES>} \\
    & \texttt{<ES>} \textit{QueryAttr(}\texttt{<K>}\textit{)} & For \texttt{<ES>}, what is [ his/her | its ] \texttt{<K>} \\
    & \texttt{<ES>} \texttt{<ES>} \textit{QueryRelation()} & What is the relation from \texttt{<ES>} to \texttt{<ES>} \\
    & \texttt{<ES>} \textit{SelectAmong(}\texttt{<K>},\texttt{<SOP>}\textit{)} & Among \texttt{<ES>}, which one has the \texttt{<SOP>} \texttt{<K>}  \\
    & \texttt{<ES>} \texttt{<ES>} \textit{SelectBetween(}\texttt{<K>},\texttt{<COP>}\textit{)} & Which one has the \texttt{<COP>} \texttt{<K>}, \texttt{<ES>} or \texttt{<ES>} \\
    & \texttt{<ES>} [ \textit{QueryAttr(}\texttt{<K>}\textit{)} | \textit{QueryAttrUnderCondition(}\texttt{<K>},\texttt{<QK>},\texttt{<QV>}\textit{)} ] \texttt{<Verify>} & For \texttt{<ES>}, is \texttt{<K>} \texttt{<Verify>} [(\texttt{<QK>} is \texttt{<QV>})]? \\
    & \texttt{<ES>} \textit{QueryAttrQualifier(}\texttt{<K>},\texttt{<V>},\texttt{<QK>}\textit{)} & For \texttt{<ES>}, [ his/her | its ] \texttt{<K>} is \texttt{<V>}, [ What | Who ] is the \texttt{<QK>} \\
    & \texttt{<ES>} \texttt{<ES>} \textit{QueryRelationQualifier(}\texttt{<P>},\texttt{<QK>}\textit{)} & \texttt{<ES>} \texttt{<P>} \texttt{<ES>}, [ What | Who ] is the \texttt{<QK>} \\
    \midrule
    \texttt{<ES>} $\rightarrow$ & \texttt{<ES>} \texttt{<ES>} \texttt{<Bool>} & \texttt{<ES>} \texttt{<Bool>} \texttt{<ES>} \\
    & \texttt{<E>} & \texttt{<E>} \\
    \midrule
    \texttt{<E>} $\rightarrow$ & [ \texttt{<ES>} | \textit{FindAll()} ] \texttt{<FilterAttr>} \texttt{<FilterQualifier>}? \textit{FilterConcept(}\texttt{<C>}\textit{)}? & the [ one | \texttt{<C>} ] whose \texttt{<FilterAttr>} \texttt{<FilterQualifier>}? \\
    & [ \texttt{<ES>} | \textit{FindAll()} ] \textit{Relate(}\texttt{<P>}, DIR\textit{)} \texttt{<FilterQualifier>}? \textit{FilterConcept(}\texttt{<C>}\textit{)}? & the [ one | \texttt{<C>} ] that \texttt{<P>} \texttt{<ES>} \texttt{<FilterQualifier>}?\\
    & \textit{FindAll()} \textit{FilterConcept(}\texttt{<C>}\textit{)} & \texttt{<C>} \\
    & \textit{Find(Name)} & \textit{Name}
    \\
    \midrule
    \texttt{<FilterAttr>} $\rightarrow$ & \textit{FilterStr(}\texttt{<K>},\texttt{<V>}\textit{)} & \texttt{<K>} is \texttt{<V>} \\ 
    & \textit{FilterNum(}\texttt{<K>},\texttt{<V>},\texttt{<OP>}\textit{)} & \texttt{<K>} is \texttt{<OP>} \texttt{<V>} \\
    & \textit{FilterYear(}\texttt{<K>},\texttt{<V>},\texttt{<OP>}\textit{)} & \texttt{<K>} is \texttt{<OP>} \texttt{<V>} \\
    & \textit{FilterDate(}\texttt{<K>},\texttt{<V>},\texttt{<OP>}\textit{)} & \texttt{<K>} is \texttt{<OP>} \texttt{<V>} \\
    \midrule
    \texttt{<FilterQualifier>} $\rightarrow$ & \textit{QFilterStr(}\texttt{<QK>},\texttt{<QV>}\textit{)} & (\texttt{<QK>} is \texttt{<QV>}) \\
    & \textit{QFilterNum(}\texttt{<QK>},\texttt{<QV>},\texttt{<OP>}\textit{)} & (\texttt{<QK>} is \texttt{<OP>} \texttt{<QV>}) \\
    & \textit{QFilterYear(}\texttt{<QK>},\texttt{<QV>},\texttt{<OP>}\textit{)} & (\texttt{<QK>} is \texttt{<OP>} \texttt{<QV>}) \\
    & \textit{QFilterDate(}\texttt{<QK>},\texttt{<QV>},\texttt{<OP>}\textit{)} & (\texttt{<QK>} is \texttt{<OP>} \texttt{<QV>}) \\
    \midrule
    \texttt{<Verify>} $\rightarrow$ & \textit{VerifyStr(}\texttt{<V>}\textit{)} &  \texttt{<V>} \\ 
    & \textit{VerifyNum(}\texttt{<V>},\texttt{<OP>}\textit{)} & \texttt{<OP>} \texttt{<V>} \\
    & \textit{VerifyYear(}\texttt{<V>},\texttt{<OP>}\textit{)} & \texttt{<OP>} \texttt{<V>} \\
    & \textit{VerifyDate(}\texttt{<V>},\texttt{<OP>}\textit{)} & \texttt{<OP>} \texttt{<V>} \\
    \midrule
    \texttt{<Bool>} $\rightarrow$ & \textit{And()} & and \\
    & \textit{Or()} & or \\
    \midrule
    \texttt{<SOP>} $\rightarrow$ & largest | smallest & largest | smallest \\
    \texttt{<COP>} $\rightarrow$ & greater | less & greater | less \\
    \texttt{<OP>} $\rightarrow$ & = | != | < | > & [ equal to | in | on ] | [ not equal to | not in ] | [ less than | before ] | [ greater than | after ] \\
    \midrule
    \texttt{<K>}/\texttt{<QK>} $\rightarrow$ & \textit{Key | QKey} & \textit{Key\_Text | QKey\_Text} \\
    \texttt{<V>}/\texttt{<QV>} $\rightarrow$ & \textit{Value | QValue} & \textit{Value | QValue} \\
    \texttt{<C>} $\rightarrow$ & \textit{Concept} & \textit{Concept} \\
    \texttt{<P>}$\rightarrow$ & \textit{Pred} & \textit{Pred\_Text} \\
	\bottomrule

\end{tabular}
} 
\caption{SCFG rules for producing KoPL program and canonical question pairs. ``$|$'' matches either expression in a group. ``$?$'' denotes the expression preceding it is optional. \textit{Key\_Text}, \textit{QKey\_Text}, and \textit{Pred\_Text} denote the annotated template for attribute keys, qualifier keys, and relations. For example, for \textit{Pred} \textit{place of birth}, the \textit{Pred\_Text} is ``was born in''.}
\label{tab:scfg}
\end{table*}

\section{Grammar Rules of KoPL}
\label{apdx:grammar}
As shown in Table~\ref{tab:scfg}, the supported program space of KoPL can be defined by a synchrounous context-free grammar (SCFG), which is widely used to generate logical forms paired
with canonical questions \cite{overnight, Jia2016DataRF, Wu2021FromPT}. The programs are meant to cover the desired set of compositional functions, and the canonical questions are meant to capture the meaning of the programs. 


\section{Knowledge Base Extraction}
\label{apdx:KB extraction}
Specifically, we took the entities of FB15k-237~\cite{FB15K-237}, a popular subset of Freebase, as seeds, and then aligned them with Wikidata via Freebase IDs\footnote{Wikidata provides the Freebase ID for most of its entities, but the relations are not aligned.}, so that we could extract their rich literal and qualifier knowledge from Wikidata.
Besides, we added 3,000 other entities with the same name as one of FB15k-237 entities, to further increase the difficulty of disambiguation.
For the relational knowledge, we manually merged the relations of FB15k-237 (\textit{e.g.}, \textit{/people/person/spouse\_s./people/marriage/spouse}) and Wikidata (\textit{e.g.}, \textit{spouse}), obtaining 363 relations totally.
Finally, we manually filtered out useless attributes (\textit{e.g.}, about images and Wikidata pages) and entities (\textit{i.e.}, never used in triples).

\section{Generation Strategies}
\label{apdx:generation strategies}
\begin{table*}[ht]
  \centering
  \scriptsize
  \begin{tabularx}{\textwidth}{l *{3}{Y}}
  \toprule
  \textbf{Strategy}  &  \textbf{Template}  &  \textbf{Example} \\
  \midrule
  \multicolumn{3}{c}{\textbf{Locating Stage}} \\
  \midrule
  Entity Name  &   -   &   LeBron James \\
  \cmidrule(lr){1-3}
  Concept Name  &  \texttt{<C>}  &  basketball players \\
  \cmidrule(lr){1-3}
  Concept + Literal  &   the \texttt(<C>) whose \texttt{<K>} is \texttt{<OP>} \texttt{<V>} (\texttt{<QK>} is \texttt{<QV>})  &   the basketball team whose social media followers is greater than 3,000,000 (point in time is 2021) \\
  \cmidrule(lr){1-3}
  Concept + Relational  &  the \texttt(<C>) that \texttt{<P>} \texttt{<E>} (\texttt{<QK>} is \texttt{<QV>})  &   the basketball player that was drafted by Cleveland Cavaliers \\
  \cmidrule(lr){1-3}
  Recursive Multi-Hop  &   unfold \texttt{<E>} in a \textit{Concept + Relational} description   &   the basketball player that was drafted by the basketball team whose social media followers is greater than 3,000,000 (point in time is 2021) \\
  \cmidrule(lr){1-3}
  Intersection   &  Condition 1 \textit{and} Condition 2  &   the basketball players whose height is greater than 190 centimetres and less than 220 centimetres \\
  \cmidrule(lr){1-3}
  Union   &   Condition 1 \textit{or} Condition 2   &   the basketball players that were born in Akron or Cleveland \\
  \midrule
  \multicolumn{3}{c}{\textbf{Asking Stage}} \\
  \midrule
  QueryName    &    What/Who is \texttt{<E>}    &    Who is the basketball player whose height is equal to 206 centimetres? \\
  \cmidrule(lr){1-3}
  Count    &    How many \texttt{<E>}     &     How many basketball players that were drafted by Cleveland Cavaliers? \\
  \cmidrule(lr){1-3}
  QueryAttribute   &   For \texttt{<E>}, what is his/her/its \texttt{<K>} (\texttt{<QK>} is \texttt{<QV>})   &  For Cleveland Cavaliers, what is its social media follower number (point in time is January 2021)?  \\
  \cmidrule(lr){1-3}
  Relation   &   What is the relation from \texttt{<E>} to \texttt{<E>}   &   What is the relation from LeBron James Jr. to LeBron James? \\
  \cmidrule(lr){1-3}
  SelectAmong    &    Among \texttt{<E>}, which one has the largest/smallest \texttt{<K>}    &   Among basketball players, which one has the largest mass? \\
  \cmidrule(lr){1-3}
  SelectBetween    &    Which one has the larger/smaller \texttt{<K>}, \texttt{<E>} or \texttt{<E>}   &   Which one has the larger mass, LeBron James Jr. or LeBron James?  \\
  \cmidrule(lr){1-3}
  Verify   &   For \texttt{<E>}, is his/her/its \texttt{<K>} \texttt{<OP>} \texttt{<V>} (\texttt{<QK>} is \texttt{<QV>})   &  For the human that is the father of LeBron James Jr., is his/her height greater than 180 centimetres? \\
  \cmidrule(lr){1-3}
  QualifierLiteral   &   For \texttt{<E>}, his/her/its \texttt{<K>} is \texttt{<V>}, what is the \texttt{<QK>}   &   For Akron, its population is 199,110, what is the point in time? \\
  \cmidrule(lr){1-3}
  QualifierRelational   &   \texttt{<E>} \texttt{<P>} \texttt{<E>}, what is the \texttt{<QK>}   &   LeBron James was drafted by Cleveland Cavaliers, what is the point in time?  \\
  \bottomrule
  \end{tabularx}
  \caption{\label{tab:templates} Templates and examples of our locating and asking stage. Placeholders in template have specific implication: \texttt{<E>}-description of an entity or entity set; \texttt{<C>}-concept; \texttt{<K>}-attribute key; \texttt{<OP>}-operator, selected from \{=, !=, $<$, $>$\}; \texttt{<V>}-attribute value; \texttt{<QK>}-qualifier key; \texttt{<QV>}-qualifier value; \texttt{<P>}-relation description, \textit{e.g.}, \textit{was drafted by}.}
\end{table*}

Table~\ref{tab:templates} list the complete generation strategies, including 7 locating and 9 asking strategies.
In locating stage we describe a single entity or an entity set with various restrictions, while in asking stage we query specific information about the target entity or entity set.

\section{SPARQL Implementation Details}
\label{apdx:sparql}
We build a SPARQL engine with Virtuoso~\footnote{https://github.com/openlink/virtuoso-opensource} to execute generated SPARQLs.
To denote qualifiers, we create a virtual node for those literal and relational triples.
For example, to denote the point in time of \textit{(LeBron James, drafted by, Cleveland Cavaliers)}, we create a node \textit{\_BN} which connects to the subject, the relation, and the object with three special edges, and then add \textit{(\_BN, point in time, 2003)} into the graph.
Similarly, we use virtual node to represent the attribute value of number type, which has an extra unit.
For example, to represent the height of \textit{LeBron James}, we need \textit{(LeBron James, height, \_BN), (\_BN, value, 206), (\_BN, unit, centimetre)}.

\section{Generation Examples}
\label{apdx:generation examples}
\begin{figure*}[ht]
  \centering
  \includegraphics[width=\linewidth]{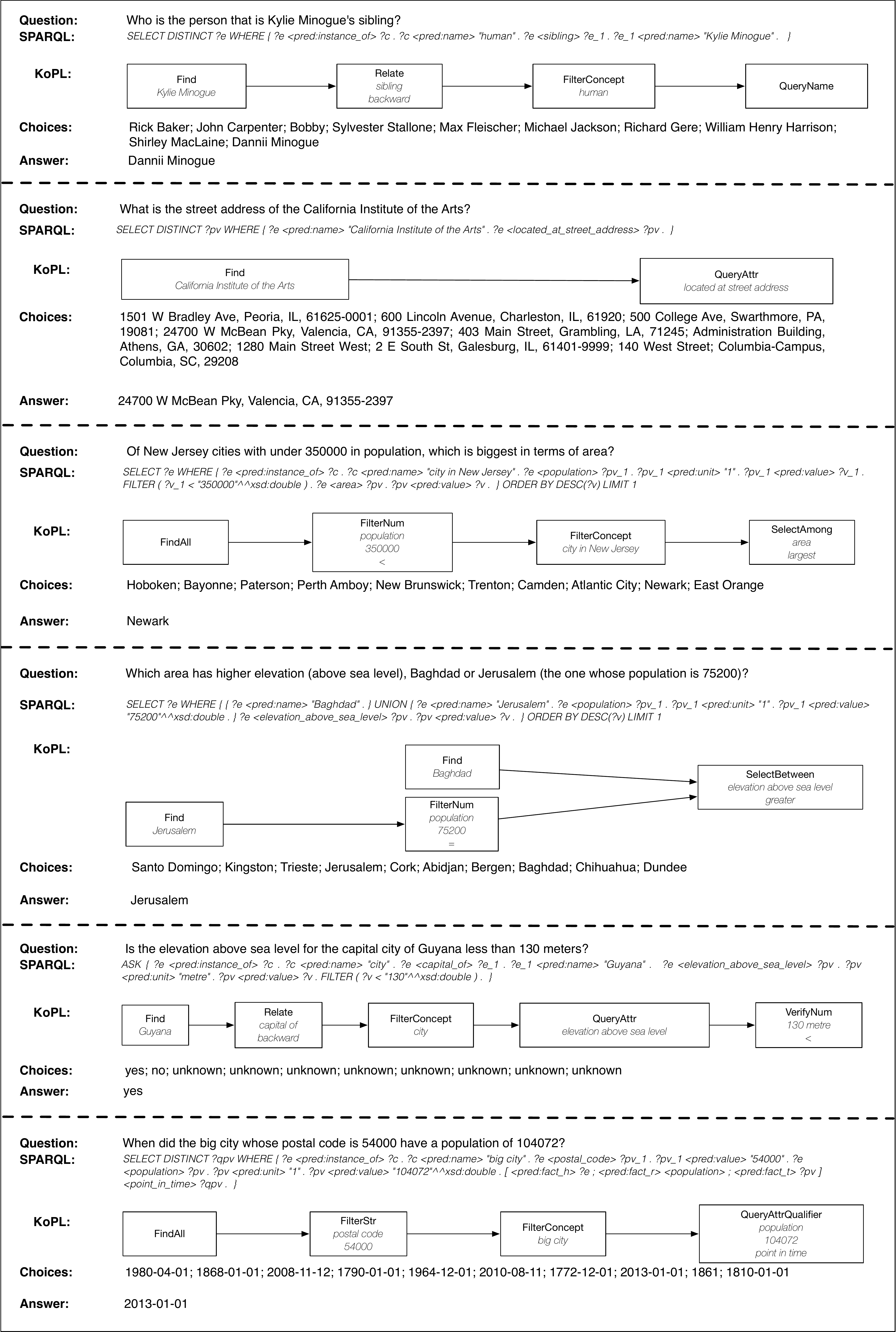}
  \caption{Examples of KQA Pro. In KQA Pro, each instance consists of 5 components: the textual question, the corresponding SPARQL, the corresponding KoPL, 10 candidate choices, and the golden answer. Choices are separated by semicolons in this figure. For questions of \emph{Verify} type, the choices are composed of ``yes'', ``no'', and 8 special token ``unknown'' for padding.}
  \label{fig:data_examples1}
\end{figure*}

Consider the example of Fig.~\ref{fig:generation} in Section~\ref{sec:strategy} in the main text, following is a detailed explanation.
At the first, the asking stage samples the strategy \textit{QueryName} and samples \textit{Cleveland Cavaliers} from the whole entity set as the target entity.
The corresponding textual description, SPARQL, and KoPL of this stage is ``\textit{Who is $<$E$>$}'', ``\textit{SELECT ?e WHERE \{ \}}'', and ``\textit{QueryName}'', respectively.
Then we switch to the locating stage to describe the target entity \textit{Cleveland Cavaliers}.
We sample the strategy, \textit{Concept + Relational}, to locate it.
For the concept part, we sample \textit{team} from all concepts of \textit{Cleveland Cavaliers}.
The corresponding textual description, SPARQL, and KoPL is ``\textit{team}'', ``\textit{?e $<$pred:instance\_of$>$ ?c . ?c $<$pred:name$>$ “team” .}'', and ``\textit{FilterConcept(team)}'', respectively.
For the relation part, we sample \textit{(LeBron James, drafted by)} from all triples of \textit{Cleveland Cavaliers}.
The corresponding textual description, SPARQL, and KoPL is ``\textit{drafted LeBron James}'', ``\textit{?e\_1 $<$drafted by$>$ ?e . ?e\_1 $<$pred:name$>$ 'LeBron James' .}'', ``\textit{Find(LeBron James) $\,\to\,$ Relate(drafted by, backward)}'', respectively.
The locating stage combines the concept and the relation, obtaining the entity description ``\textit{the team that drafted LeBron James}'' and the corresponding SPARQL and KoPL.
Finally, we combine the results of the two stages and output the complete question.
Figure~\ref{fig:data_examples1} and~\ref{fig:data_examples2} show more examples of KQA Pro.

\section{Data Analysis}
\label{sec:anaylisis in appendix}
\begin{figure}[ht]
  \centering
  \includegraphics[width=1.0\linewidth]{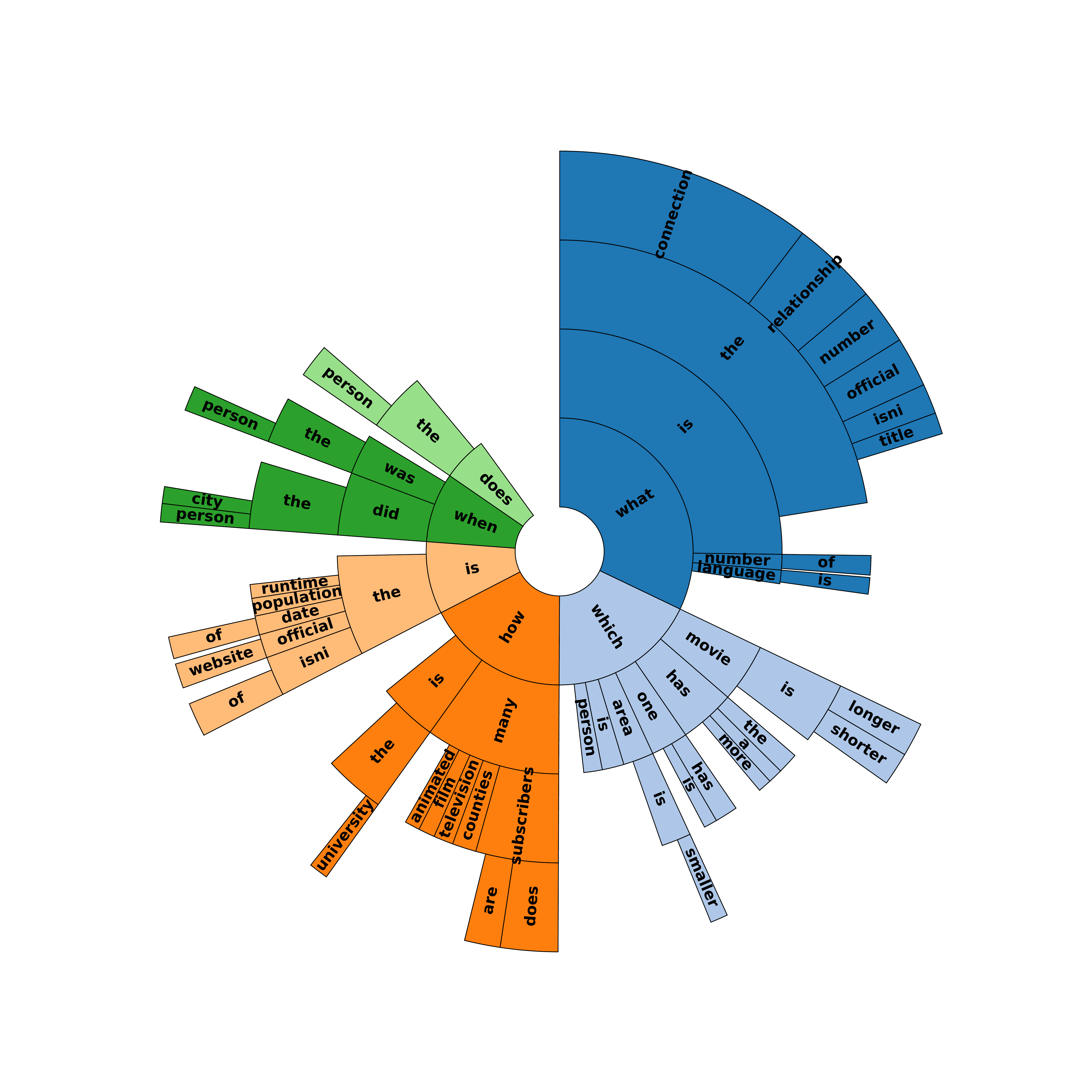}
  \caption{Distribution of first 4 question words.}
  \label{fig:sunburst}
\end{figure}


There are 24,724 unique answers in KQA Pro.
We show the top 20 most frequent answers and their fractions in Fig.~\ref{fig:answerstatistics}.
``yes'' and ``no'' are the most frequent answers, because they cover all questions of type \textit{Verify}.
``0'', ``1'', ``2'', ``3'', and other quantity answers are for questions of type \textit{Count}, which accounts for 11.5\%.

Fig.~\ref{fig:programlenstatistics} shows the Program length distribution.
Most of our problems (28.42\%) can be solved by 4 functional steps.
Some extreme complicated ones (1.24\%) need more than 10 steps.

Fig.~\ref{fig:sunburst} shows sunburst for first 4 words in questions.
We can see that questions usually start with ``what'', ``which'', ``how many'', ``when'', ``is'' and ``does''.
Frequent topics include ``person'', ``movie'', ``country'', ``university'', and etc.

\section{Baseline Implementation Details}


\noindent \textbf{KVMemNet}. 
For literal and relational knowledge, we concatenated the subject and the attribute/relation as the memory key, \textit{e.g.}, ``LeBron James drafted by'', leaving the object as the memory value.
For high-level knowledge, we concatenated the fact and the qualifier key as the memory key, \textit{e.g.}, ``LeBron James drafted by Cleveland Cavaliers point in time''.
For each question, we pre-selected a small subset of the KB as its relavant memory. 
Following \cite{miller2016key}, we retrieved 1,000 key-value pairs where the key shares at least one word with the question with frequency $<$ 1000 (to ignore stop words).
KVMemNet iteratively updates a query vector by reading the memory attentively.
In our experiment we set the update steps to 3.

\noindent \textbf{SRN}. 
SRN can only handle relational knowledge. 
It must start from a topic entity and terminate with a predicted entity. 
So we filtered out questions that contain literal knowledge or qualifier knowledge, retaining 5,004 and 649 questions as its training set and test set.
Specifically, we retained the questions with \textit{Find} as the first function and \textit{QueryName} as the last function. 
The textual input of the first \textit{Find} was regarded as the topic entity and was fed into the model during both training and testing phase.

\noindent \textbf{EmbedKGQA}. 
EmbedKGQA utilizes knowledge graph embedding to improve multi-hop reasoning.
To adapt to existing knowledge embedding techniques, we added virtual nodes to represent the qualifier knowledge of KQA Pro.
Different from SRN, we applied EmbedKGQA on the entire KQA Pro dataset, because its classification layer is more flexible than SRN and can predict answers outside the entity set.
The topic entity of each question was extracted from the golden program and then fed into the model during both training and testing.

\noindent \textbf{RGCN}.
To build the graph, we took entities as nodes, connections between them as edges, and relations as edge labels.
We concatenated the literal attributes of an entity into a sequence as the node description.
For simplicity, we ignored the qualifier knowledge.
Given a question, we first initialized node vectors by fusing the information of node descriptions and the question, then conducted RGCN to update the node features, and finally aggregated features of nodes and the question to predict the answer via a classification layer.
Our RGCN implementation is based on DGL,\footnote{https://github.com/dmlc/dgl} a high performance Python package for deep learning on graphs. 
Due to the memory limit, we set the graph layer to 1 and set the hidden dimension of nodes and edges to 32.

\noindent \textbf{RNN-based KoPL and SPARQL Parsers}. 
For KoPL prediction, we first parsed the question to the sequence of functions, and then predicted textual inputs for each function.
We used Gated Recurrent Unit (GRU)~\cite{cho2014GRU,chung2014gru}, a well-known variant of RNNs, as our encoder of questions and decoder of functions.
Attention mechanism~\cite{dong2016language} was applied by focusing on the most relavant question words when predicting each function and each textual input.
The SPARQL parser used the same encoder-decoder structure to produce SPARQL token sequences.
We tokenized the SPARQL query by delimiting spaces and some special punctuation symbols.





\section{BART KoPL Accuracy of different \#hops}
In Table~\ref{tab:hop}, we presents the BART KoPL accuracy of different \#hops. Note that
KQA Pro not only consider multi-hop relations, but also consider attributes and qualifiers. We count all of them into the hop number. So in KQA Pro, given a question with ``4-hops'', it does not mean 4 relations, but may be 1 relations + 2 attributes + 1 comparison. \textit{E.g.}, ``Who is taller, LeBron James Jr. or his father?''.

\begin{table}[htbp]
    \scriptsize
	\centering
	\small
		\begin{tabular}{cccc}
			\toprule
			2-3 Hop & 4-5 Hop & 6-7 Hop & 8-9 Hop \\
			\midrule
			92.4 	& 87.71 & 86.41 & 86.51 \\
			\bottomrule
		\end{tabular}%
    \caption{BART KoPL Accuracy of different \#hops.}
    \label{tab:hop}
\end{table}%



\begin{figure*}[ht]
  \centering
  \includegraphics[width=\linewidth]{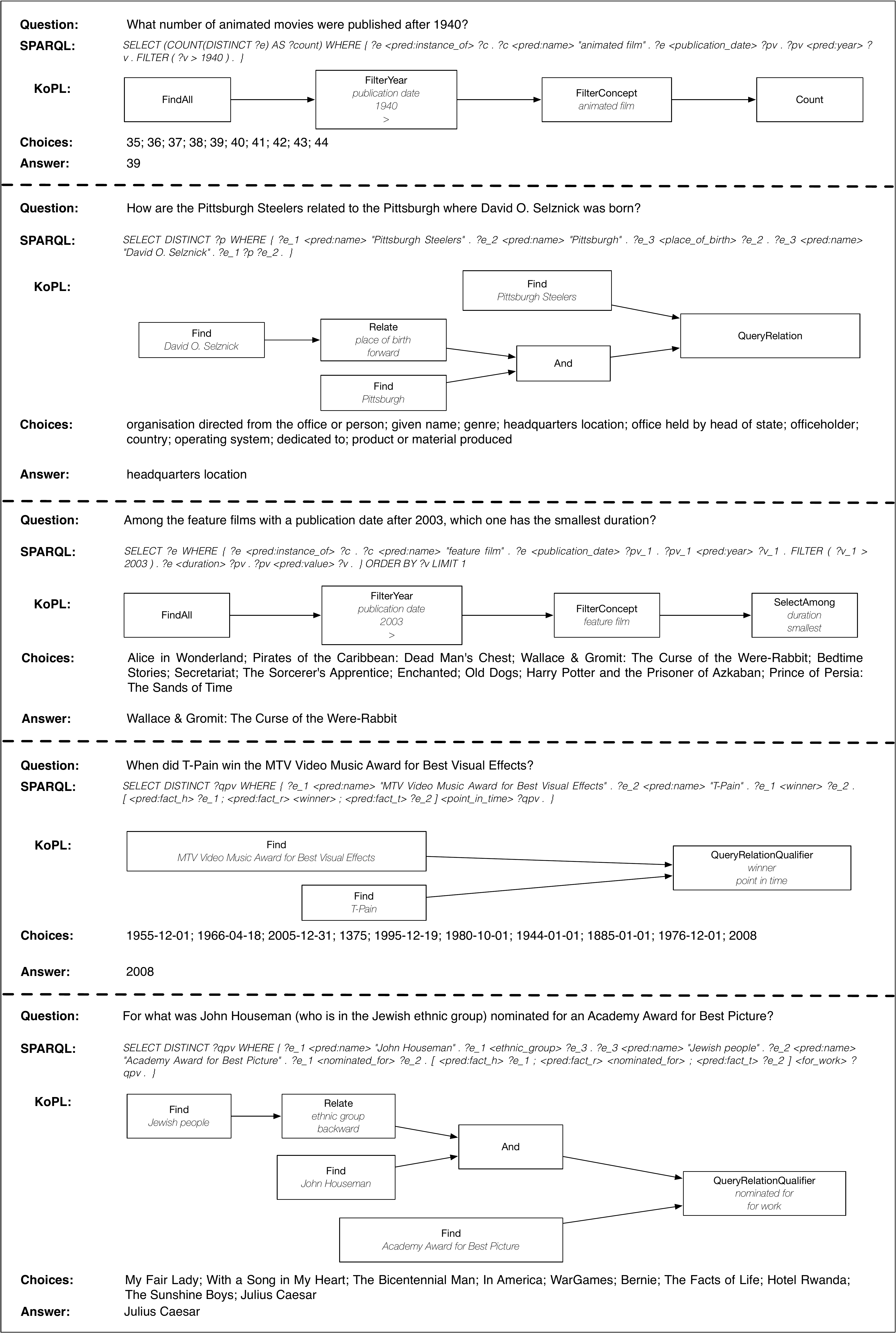}
  \caption{Examples of KQA Pro. }
  \label{fig:data_examples2}
\end{figure*}


\end{document}